\documentclass{article}
\usepackage{spconf}
\usepackage{setspace}
\usepackage[numbers]{natbib}
\usepackage{ulem}
\usepackage{amsmath}
\usepackage{amsthm}
\usepackage{graphicx}
\usepackage{subfigure}
\usepackage{color}

\usepackage{multirow}
\usepackage{booktabs}
\usepackage{threeparttable}
\usepackage[lined,boxed,commentsnumbered, ruled]{algorithm2e}

\theoremstyle{remark}

\title{Single Image Dehazing via Model-based Deep-Learning}
\name{Zhengguo Li$^{1}$, Chaobing Zheng$^{2,*}$, Haiyan Shu$^1$, and Shiqian Wu$^{2}$}
\address{$^1$ SRO Department, Institute for Infocomm Research, 1 Fusionopolis Way, Singapore\\
$^2$ School of Information Science and Engineering, WUST, Wuhan, China}
\begin{document}
\ninept
 \maketitle
\begin{abstract}
Model-based single image dehazing algorithms restore images with sharp edges and rich details at the expense of low PSNR values. Data-driven ones  restore images with high PSNR values but with low contrast, and even some remaining haze. In this paper, a novel single image dehazing algorithm is introduced by integrating model-based and data-driven approaches. Both transmission map and atmospheric light are initialized by the model-based methods, and refined by deep learning based approaches which form a neural augmentation. Haze-free images are restored by using the transmission map and atmospheric light. Experimental results indicate that the proposed algorithm can remove haze well from real-world and synthetic hazy images.
   \end{abstract}

\section{Introduction}
Due to the effect of light scattering through small particles accumulated in the air, hazy images suffer from contrast loss of captured objects \cite{1nara2003},  color distortion \cite{He}, and reduction of dynamic range \cite{1kouf2018}. Existing works on high-level image analysis such as object detection might not perform well on the hazy images, especially for those real-world hazy images with heavy haze \cite{1LiZG2021}.  It is highly demanded to study single image dehazing.

Single image dehazing is widely studied in the fields of image processing and computer vision. Two popular types of single image dehazing algorithms are  model-based ones \cite{He,CVPR16,1LiZG2021} and data-driven ones \cite{1qin2020,1dongh2020}. The model-based ones are on top of the Koschmieders law \cite{koschmider}. They can improve the visibility of real-world hazy images well regardless of haze degree but they cannot achieve high PSNR and SSIM values on the synthetic sets of hazy images. On the other hand, the data-driven ones perform well on the synthetic sets while their performance is poor for real-world hazy images, especially for those hazy images with heavy haze \cite{1liuw2021}. It is thus desired to have a single image dehazing algorithm which is applicable for both the synthesized and the real-world hazy images.

In this paper, a novel single image dehazing algorithm is proposed by fusing the model-based and data-driven approaches.    Same as the model-based methods in \cite{He,CVPR16,1LiZG2021}, both the atmospheric light and the transmission map are required by the proposed algorithm. They are initialized by using model-based methods, and refined by data-driven approaches. The atmospheric light  is  initialized  by using the hierarchical searching method in \cite{Three} which is based on the Koschmieders law \cite{koschmider}.  The transmission map is initialized by using the dark direct attenuation prior (DDAP) in \cite{1LiZG2021} which is extended from the dark channel prior (DCP) in \cite{He}. The DDAP is applicable to all the pixels in the hazy image.  Unfortunately, the initial transmission map suffers from morphological artifacts. The morphological artifacts caused by the DDAP are then reduced by the novel haze line averaging algorithm in \cite{1LiZG2021} which is designed by using the concept of haze line in \cite{CVPR16}.

The initial atmospheric light and transmission map are then refined via  the popular generative adversarial network (GAN) \cite{1gan2014}. The initial atmospheric light and transmission map can be regarded as noisy atmospheric light and transmission map. The main function of the GAN is to reduce the noise of the initial atmospheric light and transmission map. Following this idea, the generator of the proposed GAN is constructed on top of the latest DNN for noise reduction in \cite{CycleISP} and the Res2Net in \cite{1gao2021}. The discriminator of the proposed GAN is based on the PatchGAN in \cite{1isola2017}. The proposed GAN is trained by using 500 hazy images from the multiple real-world foggy image defogging (MRFID) dataset  in  \cite{1liuw2021} and 500 hazy images from the realistic single image dehazing (RESIDE) dataset in \cite{1LiB2018}.  Existing data-driven dehazing algorithms such as \cite{1qin2020,1dongh2020,1zhao2021,1dong2020} are trained by using more than 10K hazy images in the RESIDE dataset in \cite{1LiB2018}.  All ground-truth images of haze-free images in the RESIDE dataset \cite{1LiB2018} are available. The $l_1$ loss function is applied to measure the restored images for those hazy images in the RESIDE dataset.   There is a clean image for each hazy image in the MRFID  dataset but the clean and hazy images are captured under different lighting conditions. A new loss function is  derived by using an extreme channel which is extended from the dark channel in \cite{He}. The new loss function, the adversarial loss function \cite{1gan2014}, and one more new loss function on the gradient of the restored image are used to measure the restored images for the hazy images in the MRFID dataset. The model-based estimation and the data-driven refinement forms a neural augmentation  which can be applied to improve the interpretability of pure data-driven approaches  \cite{1nir2021}.   Experimental results show that the proposed algorithm outperforms existing data-driven dehazing algorithms for the real-world hazy images from the dehazing quality index (DHQI) point of view  \cite{1min2019}.  The main contribution is to propose a novel model-based deep learning framework which is applicable to the synthetic and real-world hazy images. The number of training data  can be reduced significantly.
\begin{figure*}[htb]
	\centering
	\includegraphics[width=0.85\textwidth]{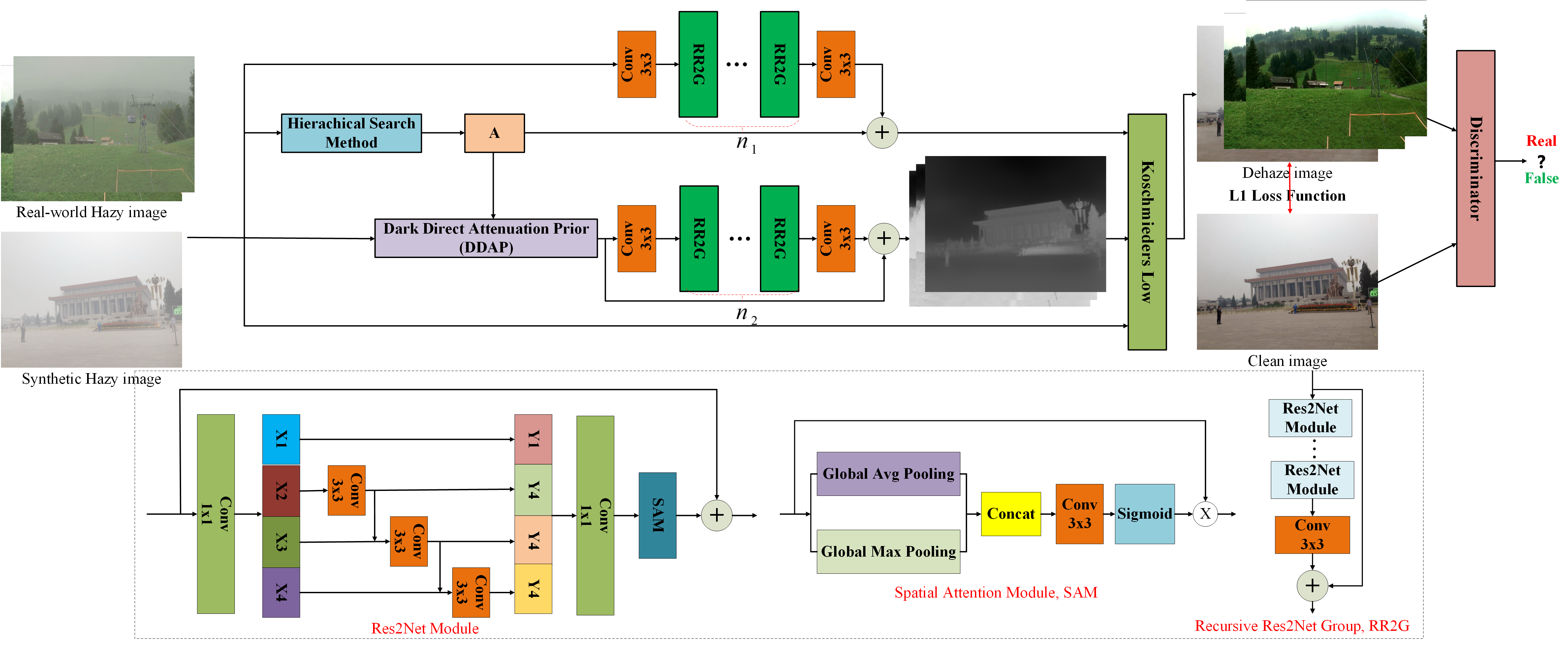}
	\caption{The proposed framework for single image dehazing via model-based GAN. The generator is on top of the DNN in \cite{CycleISP}, and the discriminator is based on the PatchGAN in \cite{1isola2017}. Each dual attention block (DAB) \cite{CycleISP} contains spatial attention and channel attention modules. The Recursive Residual Group (RRG) contains several DABs and one $3*3$ convolution layers.}
	\label{Fig3}
\end{figure*}

\section{Neural Augmented Single Image Dehazing}
\label{newalgorithm}
The atmospheric light $A$ and the transmission map $t$ are  initialized by model-based methods, and refined by data-driven approaches. The model-based and data-driven approaches form a neural augmentation \cite{1nir2021}. The overall framework is shown in Figure \ref{Fig3}.

\subsection{Model-Based Initialization of $A$ and  $t$}

In this work, the hierarchical searching method in \cite{Three} is adopted to estimate an initial value of the $A$. The hazy image $Z$ is divided into four rectangular regions. The score of each region is defined as the mean of the average pixel value subtracted by the standard deviation of the pixel values within the region along all the color channels. The region with the highest score is further divided into four smaller regions. This process is repeated until the size of the selected region is smaller than a pre-specified threshold. Within the finally selected region, the color vector, which minimizes the distance $\|[Z_r(p)-255, Z_g(p)-255, Z_b(p)-255]\|$ is selected as the atmospheric light. The initial $A$ might not be very accurate, it will be refined by using a data-driven approach.

The dark direct attenuation of a hazy image $Z$ is defined as
\begin{align}
\label{rho}
\psi_{\rho}(I*t)(p)=\min_{p'\in \Omega_{\rho}(p)}\{\min_{c\in\{R, G, B\}}\{I_c(p')t(p')\}\},
\end{align}
where $\Omega_{\rho}(p)$ is a square window centered at the pixel $p$ of a radius $\rho$ which is usually selected as 7.
By assuming that $\psi_{\rho}(I*t)(p)$ is zero, an initial transmission map $t_0(p)$ is estimated  as \cite{1LiZG2021}
\begin{align}
\label{transmap}
t_0(p)=1- \psi_{\rho}(\frac{Z}{A})(p).
\end{align}

\begin{figure*}[!htb]
\centering{
\subfigure[a hazy image]{\includegraphics[width=1.3in]{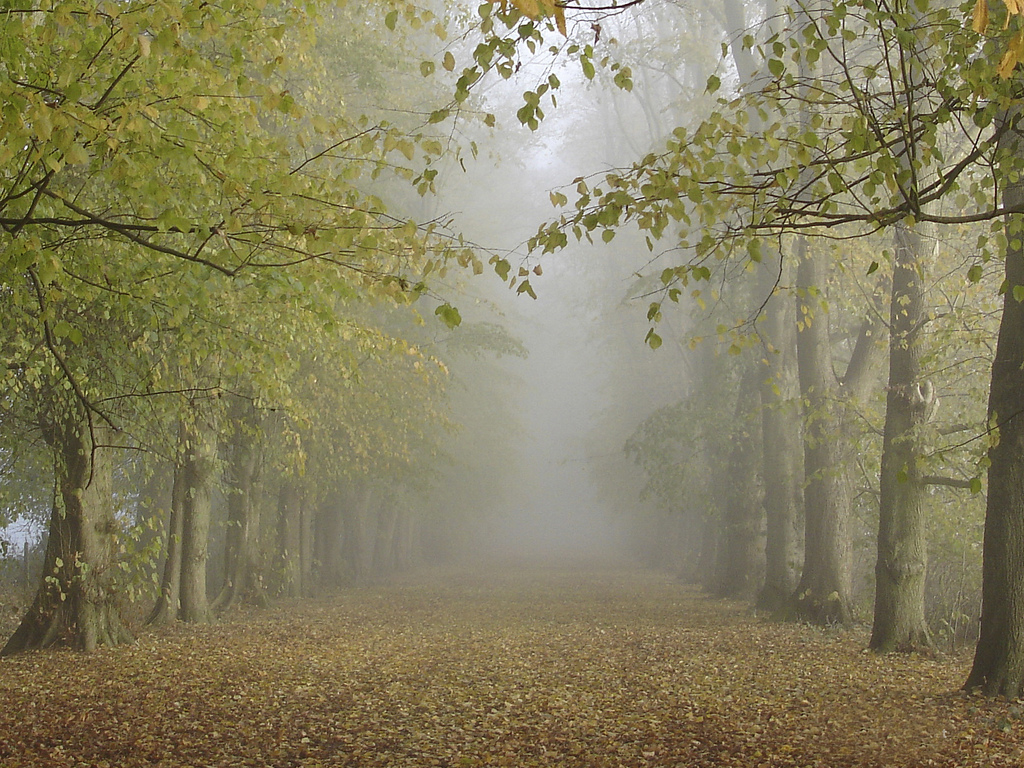}}
\subfigure[an initial transmission map]{\includegraphics[width=1.3in]{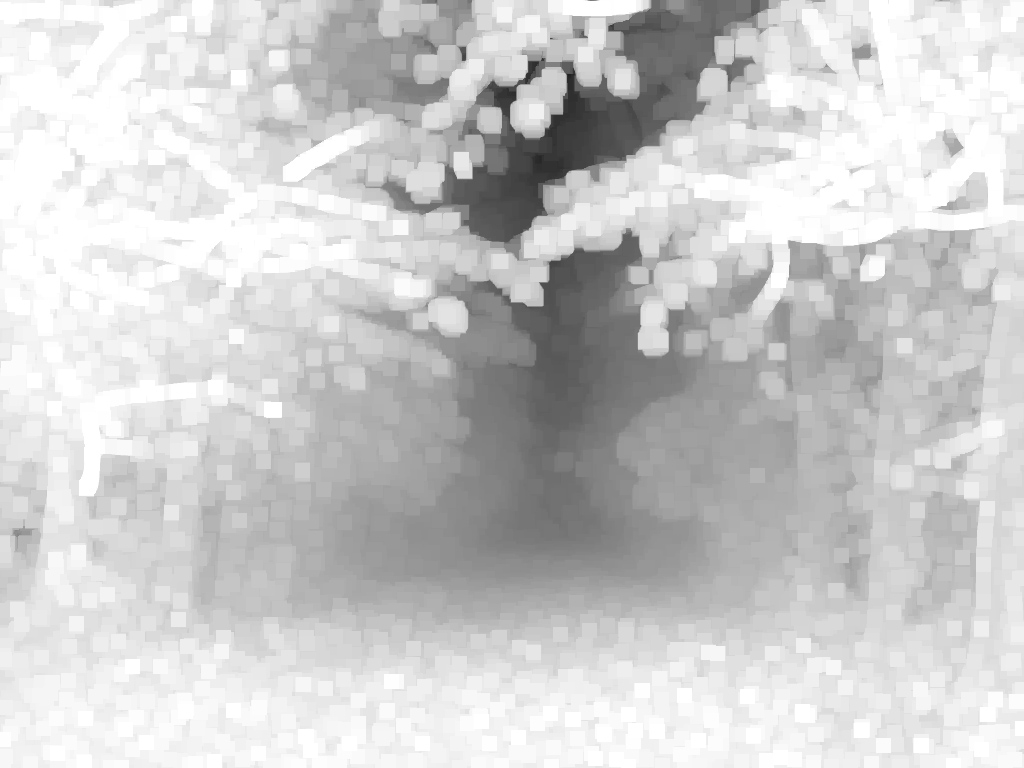}}
\subfigure[a refined transmission map by haze line averaging]{\includegraphics[width=1.3in]{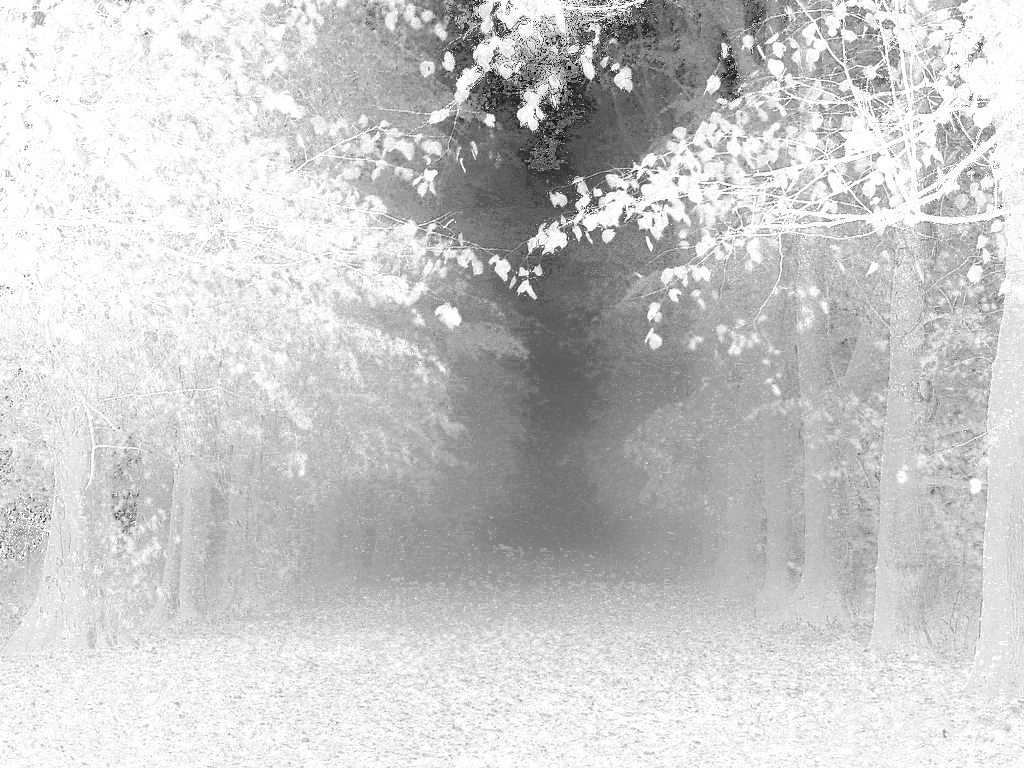}}
\subfigure[a dehazed image by initial transmission map]{\includegraphics[width=1.3in]{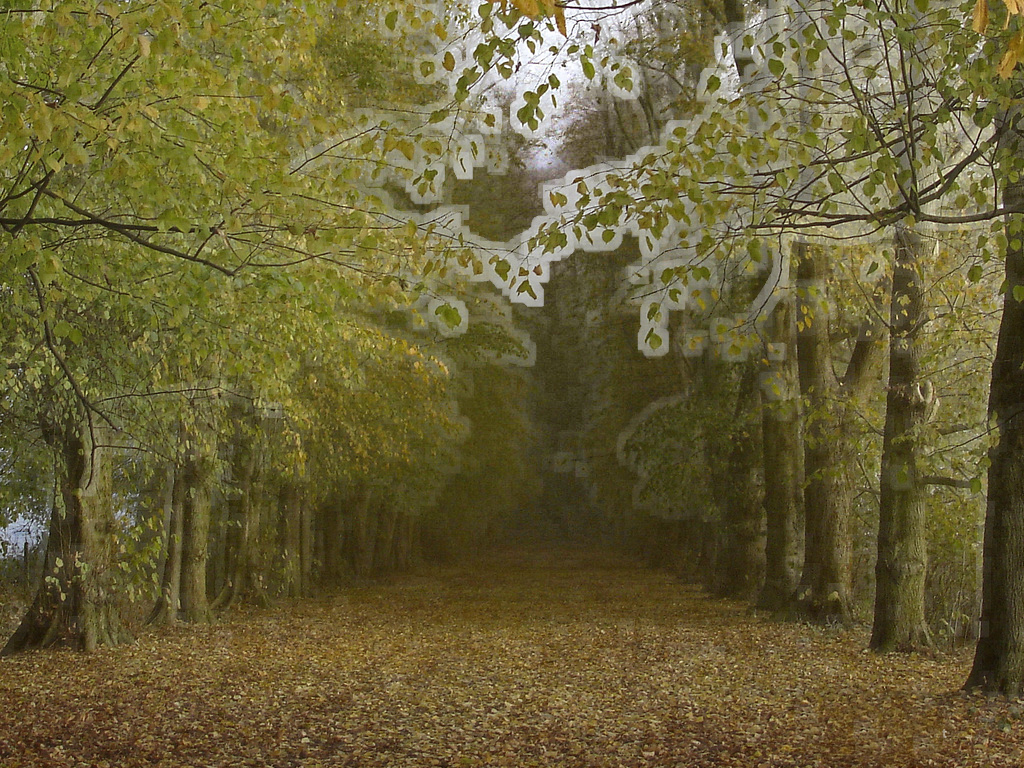}}
\subfigure[a dehazed image by refined transmission map]{\includegraphics[width=1.3in]{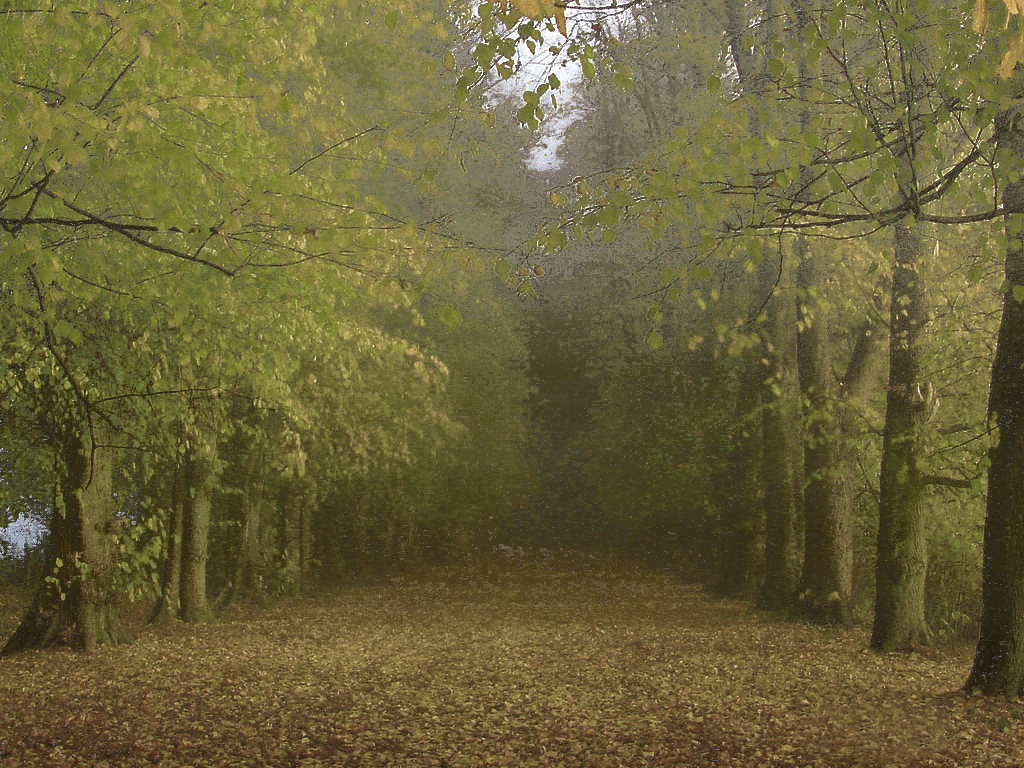}}}
\caption{Comparison of initial and refined transmission maps as well as their corresponding dehazed images.}
\label{FigFig2}
\end{figure*}

As shown in Figure \ref{FigFig2}(c), there are visibly morphological artifacts in the restored image $I$ if the $t_0(p)$ is directly applied to remove the haze from the hazy image $Z$. The haze line averaging method in \cite{1LiZG2021} is adopted to reduce the morphological artifacts. The haze line  $H(p'')$ is defined as  \cite{CVPR16}
\begin{align}
\label{hijij}
H(p'')=\{p|\sum_{c\in \{R, G,B\}}|I_c(p'')-I_c(p)|=0\}.
\end{align}

Each haze line can be identified by using a  color shift hazy pixel which is defined as $(Z(p)-A)$. The $(Z(p)-A)$ can be converted into the spherical coordinates, $[r(p)$, $\theta(p)$, $\psi(p)]$ where $\theta(p)$ and $\psi(p)$ are the longitude and latitude, respectively, and  $r(p)$ is
$\|Z(p)-A\|$. The morphological artifacts can be reduced by the following haze line averaging \cite{1LiZG2021}:
\begin{align}
\label{myrrmaxij11}
t(p)=\frac{ \sum_{p'\in H_s(p'')}t_0(p')}{\sum_{p'\in H_s(p'')}r(p')}r(p),
\end{align}
where $H_s(p'')$ is a subset of $H(p'')$, and it is obtained as follows:

A 2-D histogram binning of $\theta$ and $\psi$ with uniform edges in the range $[0,2\pi]\times [0,\pi]$ is adopted to generate an initial set of $H(p'')$. The bin size is chosen as $\pi/720\times \pi/720$. An upper bound $\nu$ is defined for the cardinality of the final sets.  It is usually selected as 200. Let $|H(p'')|$ be the cardinality of  the set $H(p'')$. The set $H(p'')$ is dividedly into $\max\{1,|H(p'')|/\nu)$ sub-sets for a non-empty $H(p'')$.

As illustrated in Figure \ref{FigFig2}(d),  the morphological artifacts are significantly reduced by the haze-line averaging (\ref{myrrmaxij11}). The remaining artifacts will be removed by a data-driven refinement.

\subsection{Data-Driven Refinement of $A$ and $t$}

Instead of using edge-preserving smoothing filters \cite{1he2013,Li2014,2kouf2015},  the GAN \cite{1gan2014} is utilized to refine the atmospheric light and transmission map. The generator is thus built up on top of the latest DNN for the noise reduction in \cite{CycleISP}. The residual net in \cite{CycleISP} is replaced by the Res2Net in \cite{1gao2021} to well capture the multi-scale features at a granular level. Both spatial attention module and  channel attention module are utilized in \cite{CycleISP}. The modules can suppress the less useful features and only allow the propagation of more informative ones. Therefore, they effectively deal with the uneven distribution of haze. The DNN for the atmospheric light is simpler than the DNN for the transmission map. The discriminator is based on the PatchGAN in \cite{1isola2017}.

The proposed GAN is trained by using 500 images from the MRFID dataset \cite{1liuw2021} and 500 images with heavy haze regenerated from the RESIDE dataset \cite{1LiB2018}. Depth is estimated by using the algorithm in \cite{1lizq2018}. The scattering coefficients of 100 images are randomly generated in $[1.2, 2.0]$ and those of the others are randomly selected in $[2.5, 3.0]$. The $A_c$'s are randomly generated in $[0.625, 1.0]$ for the color channel $c$ independently. The heavily hazy images are then generated by using the Koschmieders law \cite{koschmider}. The MRFID dataset contains foggy and clean images of 200 outdoor scenes in different lighting conditions. For each scene, one clear image and four foggy images of different densities defined as slightly, moderately, highly, and extremely foggy, are manually selected from images taken from these scenes over the course of one calendar year.

Loss functions also play an important role in the proposed GAN. Since both the atmospheric light $A$ and transmission map $t(p)$ are refined by the GAN, the loss functions are defined by using the following restored image $I$:
\begin{equation}
\label{Iij}
I(p)=\frac{Z(p)-A}{\max\{t(p),0.1\}}+A.
\end{equation}

Due to the different lighting conditions, the conventional loss functions include $l_1$ and $l_2$ loss functions are not applicable for the MRFID dataset. A new  loss function is proposed by introducing an extreme channel.  Let $\bar{I}_c(p)$ be defined as
\begin{align}
\bar{I}_c(p)=\min\{I_c(p),255-I_c(p)\},
\end{align}
and the extreme channel of the image $I$ is defined as
\begin{align}
\phi_{\rho}(I)(p)=\psi_{\rho}(\bar{I})(p).
\end{align}
Considering a pair of hazy image $Z$  and clean image $T$ which are captured from the same scene, the corresponding haze-free image of the hazy image is $I$. It can be known from the conventional imaging model \cite{1gonz2002} that
\begin{align}
[T(p), I(p)]=[L_T(p), L_I(p)]R(p),
\end{align}
where $L_T(p)$ and $L_I(p)$ are the intensities of  ambient light when the images $T$ and $I$ are captured. $R(p)$ is  the ambient reflectance coefficient of surface, and it highly depends on the smoothness or texture of the surface. Since both the $L_T(p)$ and $L_I(p)$ are  constant in a small neighborhood, $\phi_{\rho}(I)(p)$ and $\phi_{\rho}(T)(p)$ are usually determined by the reflection $R(p)$. Thus, it follows that
 \begin{align}
 \phi_{\rho}(I)(p)\approx \phi_{\rho}(T)(p) \approx 0.
 \end{align}

The extreme channel of the restored image is required to match that of the clean image $T$ while the dark channel of the restored image is required to be zero in \cite{1li2020}. Let $W$ and $H$ be the width and height of the image $I$, respectively. The first new loss function for the MRFID dataset is defined by using the extreme channels $\phi_{\rho}(I)$ and $\phi_{\rho}(T)$ as
\begin{equation}
L_e=\frac{1}{WH}\sum_p \|\phi_{\rho}(I)(p)-\phi_{\rho}(T)\|_2^2.
\end{equation}

\begin{table*}[!htb]
	\centering
	\caption{Average PSNR  value of 500 outdoor hazy images in the SOTS for different algorithms $\uparrow$.}
{\small	\begin{tabular}{c|c|c|c|c|c|c|c|c}
		\hline
 &  FD-GAN \cite{1dong2020}&  RefineDNet \cite{1zhao2021} &	FFA-Net \cite{1qin2020} & PSD \cite{1chen2021} & DCP \cite{1he2013}     & HLP \cite{CVPR16} &   MSBDN \cite{1dongh2020}  &       Ours  \\\hline
PSNR	& 20.78 & 20.80 & 32.13 & 15.15  & 17.49  & 18.06  & 30.25	  & 21.42\\
		\hline
		\end{tabular}}	
    \label{table3}
\end{table*}

\begin{table*}[htb]
	\centering
	\caption{Average DHQI values of 79 real-world outdoor hazy images for different algorithms $\uparrow$.}
{\small	\begin{tabular}{c|c|c|c|c|c|c|c|c}
\hline
 &  FD-GAN \cite{1dong2020}& RefineDNet \cite{1zhao2021} &	FFA-Net \cite{1qin2020} & PSD \cite{1chen2021} & DCP \cite{1he2013}     & HLP \cite{CVPR16} &   MSBDN \cite{1dongh2020}  &      Ours    \\\hline
 DHQI	&51.00  &57.57 &55.33 &50.60  &51.92  &52.75  &54.32  &58.88 \\
		\hline
		\end{tabular}}
    \label{table2}
\end{table*}

\begin{figure*}[!htb]
	\centering
	\includegraphics[width=0.925\textwidth]{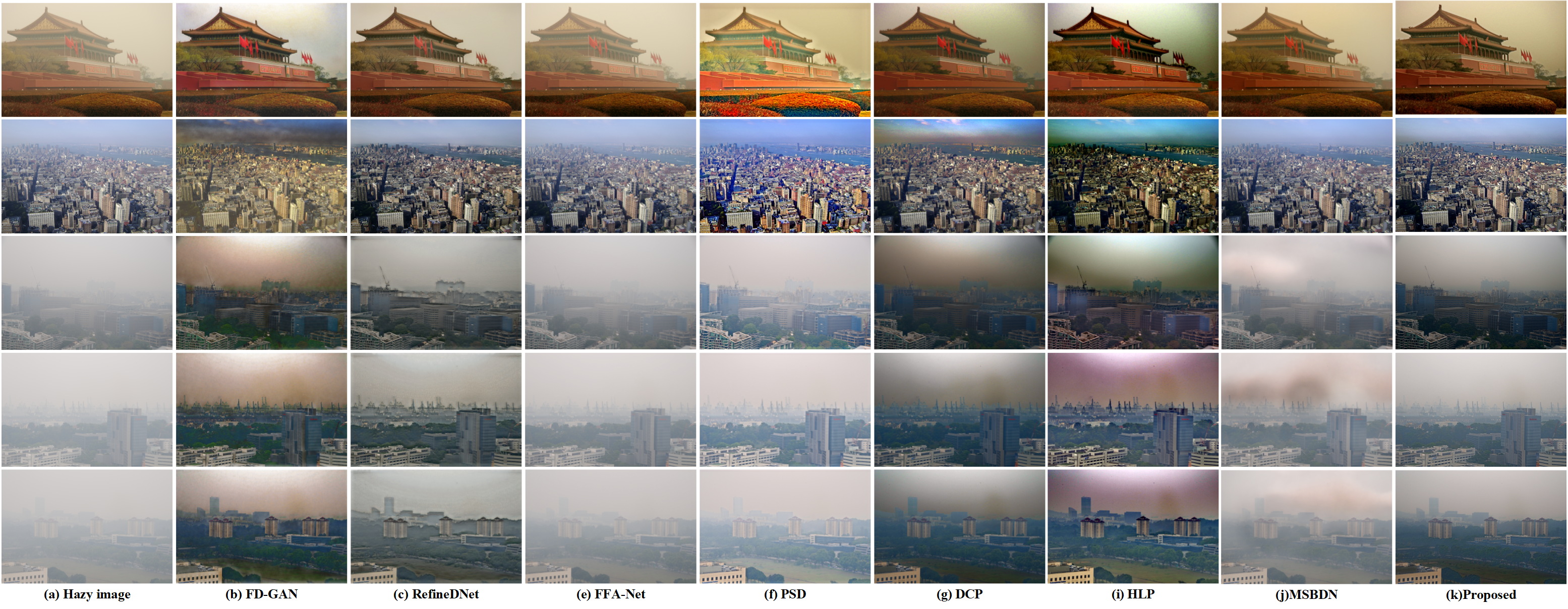}
	\caption{Comparison of different haze removal algorithms. From left to right, hazy images, dehazed images by FD-GAN \cite{1dong2020},
		RefineDNet \cite{1zhao2021}, FFA-Net \cite{1qin2020},  PSD \cite{1chen2021}, DCP \cite{1he2013}, HLP \cite{CVPR16}, MSBDN \cite{1dongh2020}, and ours, respectively. }
	\label{Fig6789}
\end{figure*}

Let $\nabla_h$ and $\nabla_v$ represent the horizontal and vertical gradients, respectively. The gradients of the restored image are required to approach those of the clean image $T$ rather than zeros as in \cite{1li2020}. The second new loss function for the MRFID dataset is defined by using the gradients of the restored image $I$  and clean image $T$ as
{\small \begin{align}
L_t = \frac{{\displaystyle \sum_{p,c}}(|\nabla_h I_c(p)-\nabla_h T_c(p)|+|\nabla_v I_c(p)-\nabla_v T_c(p)|)}{WH}.
\end{align}}

The third adversarial loss function for the MRFID dataset is defined by using the restored image $I$ and clean image $T$ as
\begin{align}
L_{adv}=\log(D(T))+\log(1-D(I)).
\end{align}

The overall loss function for the images in the MRFID dataset is defined as
\begin{equation}
\label{eq14}
L_{d,1}=  w_{adv}L_{adv}+w_e L_e +  L_t,
\end{equation}
where $w_{adv}$ and $w_e$ are two constants, and they are empirically selected as 1 and 200, respectively.

Since the ground-truth image $T$ is available in the RESIDE dataset, the loss function for the image in the RESIDE dataset is defined as
\begin{align}
L_{d,2}=\frac{1}{WH}\sum_p|I(p)-T(p)|.
\end{align}
\vspace{-5mm}
\begin{algorithm}
\label{algo3}
{
\caption{{\small Model-based single image deep dehazing}}
\begin{enumerate}
\item[Step 1.] Initialize the  atmospheric light $A_0$ from the hazy image $Z$ using the method in \cite{Three}.
\item[Step 2.] Initialize the transmission map $t_0$ by using the DDAP as in the equation (\ref{transmap}).
\item[Step 3.] Reduce the morphological artifacts of $t_0$ via the nonlocal haze line averaging (\ref{myrrmaxij11}).
\item[Step 4.] Refine the atmospheric light and transmission map using the GAN in Figure \ref{Fig3}.
\item[Step 5.] Restore the haze free image $I$ via the equation (\ref{Iij}).
\end{enumerate}}
\end{algorithm}

The atmospheric light $A$ and transmission map $t(p)$ in the equation (\ref{myrrmaxij11}) are refined by minimizing the following overall loss function in each batch:
\begin{align}
L_d=w_RL_{d,2}+L_{d,1},
\end{align}
where $w_R$ is a constant, and it is empirically selected as 100 in this paper. The proposed algorithm is summarized in algorithm \ref{algo3}.

\section{Experimental Results}
\label{experiment}
Due to the space limitation, the ablation studies are omitted and this section focuses on comparing the proposed dehazing algorithm with seven state-of-the-art ones including RefineDNet \cite{1zhao2021}, FFA-Net \cite{1qin2020}, PSD \cite{1chen2021},  DCP \cite{1he2013}, HLP \cite{CVPR16}, MSBDN \cite{1dongh2020}, and FD-GAN \cite{1dong2020}. The algorithms HLP \cite{CVPR16} and DCP \cite{1he2013} are model-based algorithms, the proposed algorithm and the RefineDNet \cite{1zhao2021} are combination of model-based and data-driven approaches,  while the others are data-driven algorithms. All the results in \cite{1zhao2021,1qin2020,1chen2021,1dongh2020,1dong2020} are generated by their publicly shared codes.

The proposed GAN is trained by using the hazy and clean images of 125 outdoor scenes from the MRFID dataset with the corresponding 500 hazy images as well as 500 images with heavy haze generated using ground-truth images from the RESIDE dataset \cite{1LiB2018}.   The hazy images of the 25 outdoor scenes from the MRFID dataset with the corresponding 100 hazy images are selected as the the validation set. All these hazy images are randomly selected from the two datasets. The test images comprise 500 outdoor hazy images in the synthetic objective testing set (SOTS) \cite{1LiB2018}, and 79 real-world hazy images in \cite{1LiZG2021}  which include 31 images are from the RESIDE \cite{1LiB2018}, and the 19 images from \cite{1fattal2008} and the Internet.

The PSNR is first adopted to compare all these algorithms by using 500 synthetic outdoor hazy images in the SOTS  \cite{1LiB2018}. The average PSNR values of the eight algorithms are given in Table \ref{table3}.  The FFA-Net \cite{1qin2020} and  MSBDN \cite{1dongh2020} are two CNN based dehazing algorithms, and they are optimized to provide high PSNR values on the SOTS dataset. Both the proposed algorithm and the algorithm in \cite{1zhao2021} indeed outperform the model-based algorithms in \cite{CVPR16} and \cite{1he2013}.

The quality index DHQI in \cite{1min2019} is then adopted to compare all these algorithms by using the 79 real-world hazy images in \cite{1LiZG2021}. The average DHQI values of the 79 real-world outdoor hazy images are given in Table \ref{table2}. The proposed algorithm outperforms others from the DHQI point of view.

Finally, all these dehazing algorithms are compared subjectively as in Figure \ref{Fig6789}.  Readers are invited to view to electronic version of figures and zoom in them so as to better appreciate differences among all the images. Although the FFA-Net \cite{1qin2020} and  MSBDN \cite{1dongh2020} achieve high PSNR and SSIM values for the synthetic hazy images, their restored results are blurry for real-world hazy images and the dehazed images are not photo-realistic. In addition, the haze is not reduced well if it is heavy. The DCP  \cite{1he2013}, HLP \cite{CVPR16}, RefineDNet \cite{1zhao2021}, FD-GAN \cite{1dong2020}, and the proposed algorithm can be applied to generate  photo realistic images.  There are visible morphological artifacts in the restored images by the  PSD \cite{1chen2021}, RefineDNet \cite{1zhao2021} and FD-GAN \cite{1dong2020}. Textures generated by the RefineDNet \cite{1zhao2021} and FD-GAN \cite{1dong2020} are different from the real ones. The DCP  \cite{1he2013} and HLP \cite{CVPR16} restore more vivid and sharper images at the expense of amplified noise in sky regions. All these problems are overcome by the proposed algorithm.

\section{Conclusion Remarks and Discussion}
\label{conclusion}

A new single image dehazing algorithm is introduced in this paper. Both transmission map and atmospheric light are obtained by a neural augmentation  which consists of model-based estimation and data-driven refinement. They are then applied to restore a haze-free image. Experimental results validate that the proposed algorithm removes haze well from  the synthetic and real-world hazy images. The proposed neural augmentation reduces the number of training data significantly. This paper focused on day-time hazy images. The proposed framework will be extended to study night-time hazy images \cite{1yangm2018} in our future research.

\end{document}